# A Combined Method of Fractal And GLCM Features for MRI and CT Scan Images Classification

Redouan Korchiyne[1, 2], Sidi Mohamed Farssi[2], Abderrahmane Sbihi[3], Rajaa Touahni[1], Mustapha Tahiri Alaoui[2]

[1] LASTID Laboratory, Ibn Tofail University- Faculty of Sciences, Kenitra, Morocco
[2] Laboratory of medical Images and Bioinformatics, Cheikh Anta Diop University-Polytechnic High School, Dakar, Senegal
[3] LTI Laboratory, Abdelmalek Essaadi University-ENSA, Tangier, Morocco

## Abstract

*Fractal analysis has been shown to be useful in image processing for characterizing shape and gray-scale complexity. The fractal feature is a compact descriptor used to give a numerical measure of the degree of irregularity of the medical images. This descriptor property does not give ownership of the local image structure. In this paper, we present a combination of this parameter based on Box Counting with GLCM Features. This powerful combination has proved good results especially in classification of medical texture from MRI and CT Scan images of trabecular bone. This method has the potential to improve clinical diagnostics tests for osteoporosis pathologies.*

## 1. Introduction

Osteoporosis is a progressive bone disease that is characterized by a decrease in bone mass and density which can lead to an increased risk of fracture [1] (see Figure1). Several methods have been applied to characterize the bone texture: (Genetic algorithm) [2] (Multi resolution analysis) [3], hybrid algorithm consisting of Artificial Neural Networks and Genetic Algorithms [4], texture analysis methods (Gabor filter, wavelet transforms and fractal dimension) [5].

Texture analysis is important in many applications of computer image analysis for classification or segmentation of images based on local spatial variations of intensity or color. A feature is an image characteristic that can capture certain visual property of the image. Four major methods were used to characterize different textures in an image: statistical methods (co-occurrence method) [6] [7], geometrical (Voroni tessellation features, fractal) [8] [9], model based method (Markov random fields) [10] [11] and signal processing methods (Gabor filters, wavelet transform and curvelets) [7] [11] [12]. The most widely used statistical methods are co-occurrence features [6] [7].

Fractal analysis has been successfully applied in images processing [13]. Applications in medical images are concerned to model the tissues and organ constitutions and analyzing different types [14] [15]. The fractal objects are characterized by: large degree of heterogeneity, self-similarity, and lack of a well-defined scale. Notion "self-similarity" means that small-scale structures of fractal set resemble large-scale structures [16] [17] [18].

Texture classification is one of the four problem domains in the field of texture analysis [19]. Texture classification based on fractal geometry proved a correlation between texture and fractal dimension [20]. In this paper, we use the combination of fractal dimension and the co-occurrence matrix features (GLCM) to describe the medical texture and used for trabecular bone texture classification.

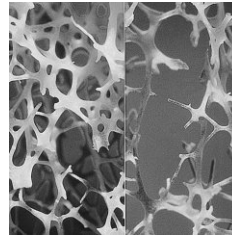

Figure 1. Left: normal bone, right: osteoporotic bone [1].

## 2. FEATURE EXTRACTION METHODS

In this work, we propose to use two common feature extraction algorithms based on global fractal descriptors and Grey Level Co-occurrences matrix (GLCM). In this section, an overview of these two algorithms is given.

Before developing the proposed feature extraction methods based on fractal and GLCM features and discussing their properties, we briefly review fractal theory and co-occurrences matrix as applicable in our work.

### 2.1. Fractal theory

Fractal geometry was introduced and supported by the mathematician Benoit Mandelbrot to characterize objects with unusual properties in classical geometry [21] [22], this concept helps to interpret the inexplicable subjects by any law. It is shown to study irregular objects in the plan or space, which is actually better suited to handle the real world. Fractal geometry has experienced significant growth in recent years, a review of this theory is summarised by R. Lopes [23]. In medical images analysis that the result of this theory has provided an evidence solutions to several problems [24] source of many applications [18] [25].

A texture can be seen as a repetition of similar patterns randomly distributed in the image. Fractal approach allows measuring the invariant translation, rotation and even scaling. Many approaches have been developed and implemented to characterize real textures using concepts of this geometry [26]. Fractal dimension is a fractional parameter and greater than the topological dimension. The fractal dimension is a helpful of an image by describing the texture irregularity [27] [28].

Practical way to obtain the fractal dimension is described as follows: a graph of $Log(N(\lambda))$ contribution to $Log(\lambda)$ for each object $\lambda$ must be established. Then the correlation between $N(\lambda)$ and $(\lambda)$ is adjusted by linear regression (see Figure. 3). Fractal Dimension is estimated using the least square method and compute the slope of the $Log$-$Log$ curve composed by the $(Log(N(\lambda)), Log(\lambda))$ points. Several methods have been introduced to calculate the fractal dimension [29] [30]. The method of boxes counting remains the most practical method to estimate the fractal dimension [31]; Box Fractal Dimension is a simplification of the Hausdorff dimension for non-strictly self-similar objects [32]. A given binary image, it is subdivided in a grid of size $MxM$ where the side of each box formed is $\lambda$ and $N(\lambda)$ represents the amount of boxes that contains one pixel, the fractal dimension is defined as follow:

$$D_f = -\lim_{\lambda \to 0} \frac{Log(N(\lambda))}{Log(\lambda)}$$

$N(\lambda)$ : Number of boxes
$\lambda$ : Length of the box

The best methods have been introduced to reduce quantization levels and it was demonstrated the utility of the fractal parameter and its variants in the transformed texture characterization images [26]. However, given that this parameter alone is not sufficient to discriminate visually different surfaces; the concepts of homogeneity of fractal dimension were examined. The Hausdorff dimension is the most studied mathematically. We are used in other work the Hausdorff multifractal spectrum to describe the different fractalities recovering the medical image [33]. It is generally the best known, but individually, it is probably the least calculated.

Table 1 shows that each shapes can be decomposed into $N$ similar copies of itself scaled by a factor of $\lambda = 0.5$. The number of half-scaled copies N equals 2, 4 and 8 for the line, square and cube respectively. This leads to fractal dimensions of $D = -Log(N(\lambda))/Log(\lambda) = 1, 2$ and $3$ as expected. However, the Sierpinski triangle, by construction, has only $N = 3$ halfscaled copies of itself leading to a non-integral fractal dimension $D = -Log(3)/Log(0.5) = 1.585$.

Table 1. Traditional geometry decomposed into $N$ similar copies of itself by a factor of $\lambda = 0.5$ and corresponding fractal dimension.

| The shapes | | $\lambda$ | $N(\lambda)$ | $D = -\dfrac{Log(N(\lambda))}{Log(\lambda)}$ |
|---|---|---|---|---|
| Line: | 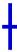 | $\lambda = \dfrac{1}{2} = 0.5$ | $N(\lambda) = 2$ | $D = -\dfrac{Log(2)}{Log(1/2)} = 1$ |
| Square: | 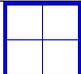 | $\lambda = \dfrac{1}{2} = 0.5$ | $N(\lambda) = 4$ | $D = -\dfrac{Log(4)}{Log(1/2)} = 2$ |
| Cube: | 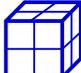 | $\lambda = \dfrac{1}{2} = 0.5$ | $N(\lambda) = 8$ | $D = -\dfrac{Log(8)}{Log(1/2)} = 3$ |
| Sierpinski Triangle: | 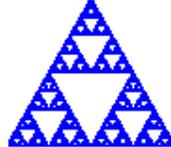 | $\lambda = \dfrac{1}{2} = 0.5$ | $N(\lambda) = 3$ | $D = -\dfrac{Log(3)}{Log(0.5)} = 1.585$ |

## 2.2. The Grey Level Co-occurrence Matrix (GLCM) Features

Grey-Level Co-occurrence Matrix (GLCM) texture measurements have been the workhorse of image texture since they were proposed by Haralick [34] [35], and 14 statistical features were introduced. GLCM is a statistical method of examining texture that considers the spatial relationship of pixels [36] [37]. The GLCM functions characterize the texture of an image by calculating how often pairs of pixel with specific values and in a specified spatial relationship occur in an image. These features are generated by calculating the features for each one of the co-occurrence matrices obtained by using the directions 0°, 45°, 90°, and 135°, then averaging these four values [38] (see Figure 3).

We illustrate this concept with a binary model and its matrix $M_{Coo}$ for d= (0°, 1). The following figure (see Figure 2) shows the calculates several values in the GLCM of the 5-by-5 image I. Element *(0,0)* in the GLCM contains the value 6 because there is 6 instance in the image where two, horizontally adjacent pixels have the values *0* and *0*. Element *(1, 1)* in the GLCM contains the value *10* because there are *10* instances in the image where two, horizontally adjacent pixels have the values 1 and 1. The *graycomatrix* function continues this processing to fill in all the values in the GLCM.

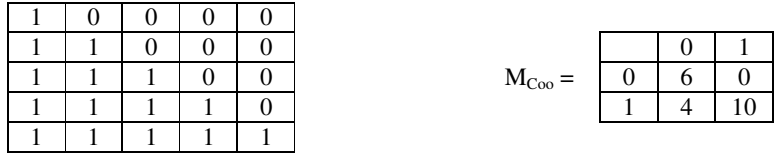

Figure 2. Binary model and corresponding co-occurrence matrix.

The some GLCM Features used in this work are:

- **Contrast**

The contrast returns a measure of the intensity contrast between a pixel and its neighbour over the whole image.

$$\sum_{i,j=0}^{N-1} P_{ij} (i-j)^2$$

Contrast is 0 for a constant image.

- **Correlation**

The correlation returns a measure of how correlated a pixel is to its neighbour over the whole image.

$$\sum_{i,j=0}^{N-1} P_{ij} \frac{(i-\mu)(j-\mu)}{\sigma^2}$$

Correlation is *1* or *-1* for a perfectly positively or negatively correlated image. Correlation is *NaN* for a constant image.

- **Energy**

The energy returns the sum of squared elements in the GLCM.
Energy is *1* for a constant image.

$$\sum_{i,j=0}^{N-1} \left( P_{ij} \right)^2$$

- **Homogeneity**

The homogeneity returns a value that measures the closeness of the distribution of elements in the GLCM to the GLCM diagonal. Homogeneity is 1 for a diagonal GLCM.

$$\sum_{i,j=0}^{N-1} \frac{P_{ij}}{1+(i-j)^2}$$

$P_{ij}$ = Element *i, j* of the normalized symmetrical GLCM.
$N$ = Number of gray levels in the image as specified by number of levels in under quantization on the GLCM.
$\mu$ = The GLCM mean, calculated as:

$$\mu = \sum_{i,j=0}^{N-1} i P_{ij}$$

$\sigma^2$ = The variance of the intensities of all reference pixel in the relationships that contributed to the GLCM, calculated as:

$$\sigma^2 = \sum_{i,j=0}^{N-1} P_{ij}\left(i - \mu\right)^2$$

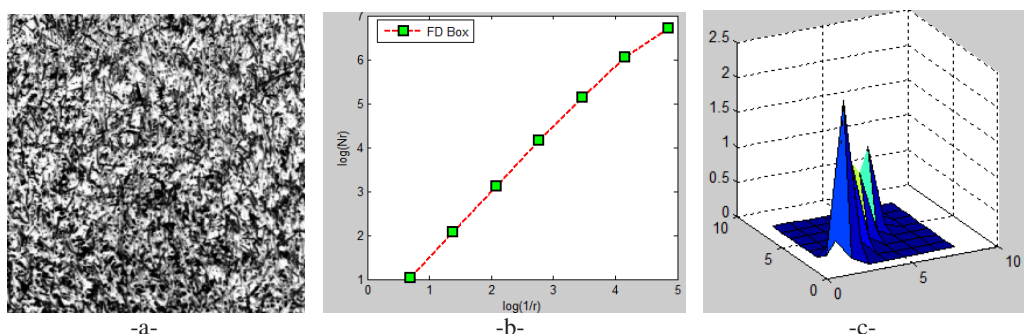

-a-  -b-  -c-

Figure 3. (a) Example from Brodatz database (b) Computed box dimension (c) GLCM Matrix.

## 3. PROPOSED METHOD

### 3.1. Principle

Several authors have proposed to analyze medical images from the texture statistics or others features as wavelets, Gabor filter, fractal dimension. Texture analysis based on the fractal concept was introduced by Pentland in 1984 [30]. In this paper, the proposed method for classification is based on the combination of fractal features and the Haralick features extracted using GLCM Matrix [36] [37].

The proposed classification approach, as shown in Figure 4, depends on maintaining the properties of Hölder transform images in minimizing the bit quota for no significant coefficients and maintaining the most significant parts of the fractal and GLCM features. Moreover, it is expected to utilize the efficiency of fractal image in detecting similarities. In the following subsections the fractal features procedure, GLCM Features, and proposed combined features scheme are further explained and clarified. This method consists of four main steps: threshold of the regions of interest, application of the Hölder image, extraction of most discriminative texture features, creation of a classifier. The images were analyzed to obtain texture parameters such as the fractal geometry based box-counting dimension, and co-occurrence parameters. Fractal dimension showed moderate correlation and decreased with GLCM features. A one tailed procedure was used to determine the differences between the texture parameters of both healthy and osteoporotic patients. The following figure (Figure 4) describes the data set classification process of our proposed method.

### 3.2. Features Extraction

#### 3.2.1. Fractal Features

The original image $I_0(i, j)$ is transformed to Hölder image $I_H(\alpha(i, j))$ and a multifractal spectrum is plotted using a parameter couple $(\alpha, f(\alpha))$ [33].

$$\alpha = \lim_{\lambda \to 0}(\alpha_k) = \lim_{\lambda \to 0} \frac{Log(\mu(S_k))}{Log(\lambda)}$$

Where $\alpha_k$ is the Hölder exponent of the subspace $S_k$.

Then, considering that each value $\alpha$ of Hölder exponent, defined a set fractal $E(\alpha)$, which will calculate the fractal dimension, and that the support of the image $I_0(i, j)$ is therefore formed the union of sets fractals of different dimensions:

$$S_k = E(\alpha_k) = \{(i, j) / \alpha(i, j) = \alpha_k\} \qquad \text{and} \quad I_H(\alpha(i, j)) = \bigcup_k E(\alpha_k)$$

The Hausdorff dimension is a way of calculating the fractal dimension based on box-counting method. This parameter is given by:

$$f_\lambda(\alpha_k) = -\frac{Log(N_\lambda(\alpha_k))}{Log(\lambda)}$$

$N_\lambda(\alpha_k)$ is the number of boxes $S_k$ containing the value $\alpha_k$

The limiting values of the multifractal spectrum $f(\alpha)$ is calculated from linear regression from a set of points in a bi-logarithmic diagram of $Log(N_\lambda(\alpha_k))$ vs $-Log(\lambda)$:

$$f(\alpha) = \lim_{\lambda \to 0}(f_\lambda(\alpha_k))$$

### 3.2.2. GLCM Features

A GLCM is a function of co-occurring greyscale values at a given offset over an image using for texture classification. We use Matlab Image Processing Toolbox to compute the four parameters (cf. Section 2.2).

In this work, we use 40 samples of two classes of different medical textures and a GLCM function with a horizontal offset of 2 and average value of four directions 0°, 45°, 90°, and 135° are computed. Next, four features of the GLCM matrices are extracted: Contrast, Correlation, Energy, and Homogeneity.

Table 2. The results of the some selected features

| Medical Images | Fractal Dimension | GLCM Features | | | | | | | |
|---|---|---|---|---|---|---|---|---|---|
| | | Contrast | | Correlation | | Energy | | Homogeneity | |
| | | Min | Max | Min | Max | Min | Max | Min | Max |
| Sample 1 | 1.9295 | 1.3243 | 1.5397 | 0.1558 | 0.2765 | 0.0924 | 0.0950 | 0.6276 | 0.6489 |
| Sample 2 | 1.8742 | 1.4691 | 1.7972 | 0.2298 | 0.3687 | 0.0665 | 0.0701 | 0.6040 | 0.6310 |

### 3.2.3. Combined Method

In the previous sections, a set of features has been analyzed for characterization of medical images and each classifier have been presented to address the osteoporosis detection. In this section, we aim to combine the information of the different classifiers so as to enhance the overall classification performance.

Fusion of information appeared to handle very large quantities of multi-source data [39]. It is to combine information from multiple sources to help in decision making. Fusion methods have been adapted and developed for applications in image processing and especially for the classification. The goal here is not to reduce redundancy in the information from several sources, but rather to consider in order improving decision making. The fusion of information depending on the architecture used here will be to combine data (most often is data or parameters from sensors or different extraction methods) for classification, whether to classify them separately according to potential of the selected classifier and then merge them.

The goal is thus to combine each output of SVM classifier and to generate a single scalar score. This will be used to make the final decision and also give information on the confidence in the decision. In this context, our research focuses on the one hand, on comparing the two methods and secondly, their combination by vote.

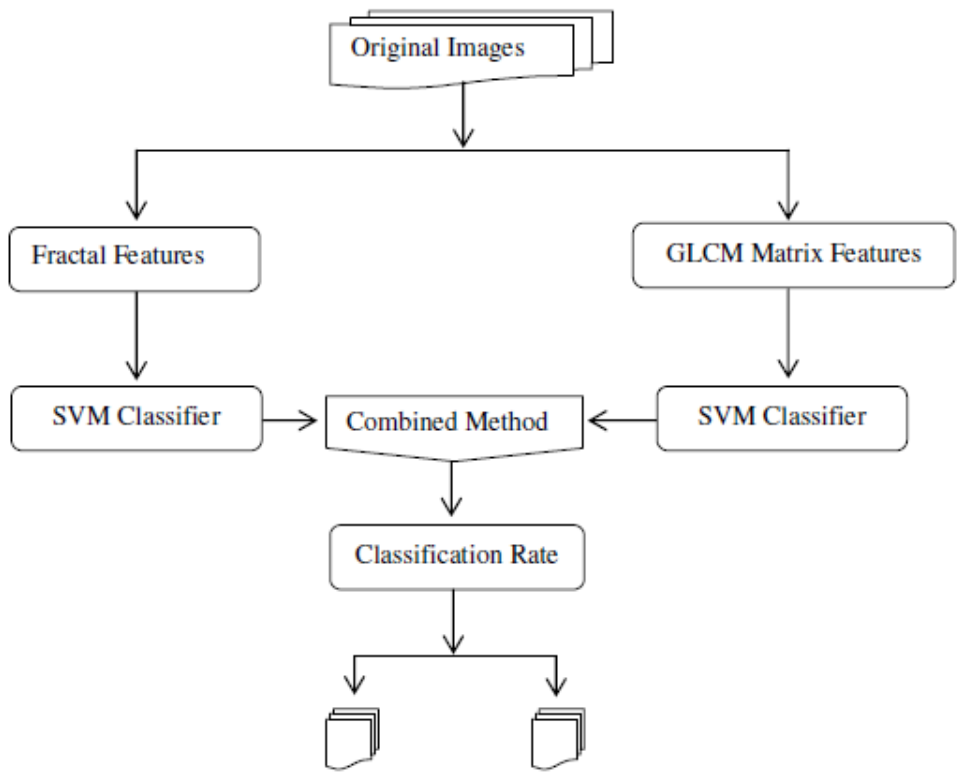

Figure 4. The flow chart of proposed method

# 4. EXPERIMENTAL RESULTS AND DISCUSSION

## 4.1. The Data Set

This paper was focused on the texture classification of medical images. The images were obtained from textures database INSERM U 703 [40], which is a set of 40 medical images from MRI (Magnetic Resonance Images) and CT scan (Computed Tomography) on a specific region namely the ROI (Region Of Interest) of trabecular bone, including 20 normal subjects and 20 abnormal subjects of osteoporosis pathology. In this work, we present the experimental details of texture classification based on fractal geometry and GLCM matrix. We used 20 ROI from MRI and CT Scan images of medical images database [40].

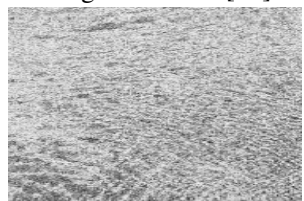 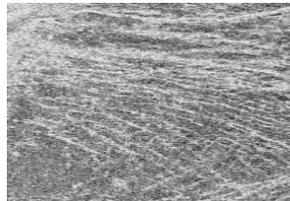

Sample1: Normal ROI        Sample 2: Pathological ROI
Figure 5. Samples from Medical Images Database [40]

### 4.2. Medical images Classification

In this section, the experiments and analyses are done to evaluate the performance of the proposed method. For each sample, a fractal features are extracted using the box counting method and 4 features are selected by the GLCM matrix [34] [36] [41]. Due to the limitation of the number of samples, we evaluate the performance method with the diverse training and test set by selecting an automatic training and test sets.

The classification was applied on 40 subjects (20 normal regions and 20 abnormal regions). The Support Vector Machine (SVM) classifier [42] [43] [44] [45], is used in this paper. The SVM [46, 47] have recently become popular as supervised classifiers of MRI data due to their high performance, their ability to deal with large high-dimensional datasets, and their flexibility in modelling diverse sources of data [48,49,50]. The most useful version is the one with the linear kernel, whereby the discriminant function between two classes is a linear combination of the inputs. The importance of a feature is associated with the absolute value of the corresponding weight in the linear combination. SVM is originally derived for two classes.The classifier was implemented in Matlab 7.8 and was investigated using a linear kernel function [22]. The performance of the classifier models were measured [23], and the parameters calculated are: (i) True Positives (TP) when the system correctly classifies subjects as normal, (ii) False Positives (FP) where the system wrongly classifies subjects, (iii) true negatives (TN) when the system correctly classifies subjects as abnormal. For the overall performance, we provide the correct classification (CCR) rate which gives the percentage of correctly classified subjects.

### 4.2.1. Statistical measures

These results will be presented as statistical measures such confusion matrix and Correct Classification Rate. In the first subsection, we define briefly these statistical measures. After that, we present the obtained results for real test of trabecular bone texture.

*Confusion Matrix:* Confusion Matrix is a 2-by-2 numeric array with diagnostic counts. This Matrix used to measure the quality of a classification system. The first column indicates the number of samples that were classified as positive, with the number of true positives in the first row, and the number of false positives in the second row. The second column indicates the number of samples that were classified as negative, with the number of false negatives in the first row, and the number of true negatives in the second row.

Correct classifications appear in the diagonal elements, and errors appear in the off-diagonal elements. Inconclusive results are considered errors and counted in the off-diagonal elements.
Table 3 shows the confusion matrix of a system that allows classifying two classes *1* and *0*. Remember that for this predictable attribute, *1* means *Normal* and *0* means *Abnormal*.

Table 3. Confusion matrix for performance evaluation.

| | | Predicted Class | |
|---|---|---|---|
| | Class | **1** | **0** |
| True Class | **1** | *TP* (True Positive) | *FN* (False Negative) |
| | **0** | *FP* (True Negative) | *TN* (True Negative) |

- *TP* (*resp. TN*) represents the number of subjects of class *1* (*resp.* class *0*) well classified by the system.
- *FN* (*resp. FP*) represents the number of subjects of class *1* (*resp.* class *0*) that have been classified by the system as subjects of class *0* (*resp.* class *1*).
- $T = TP+FN+FP+TN$ is the sum of subjects of both classes.

*Performance evaluation:* To evaluate the performance of trabecular bone classification, specificity and sensitivity used to distinguish the subjects. Sensitivity and specificity are two measures help to prove the significance of a test related to the presence or absence of pathology. Equations (1) and (2) are used to calculate these two parameters, respectively.

$$Sensitivity = \frac{TP}{TP + FN} * 100 \qquad (1)$$

$$Specificity = \frac{TN}{TN + FN} * 100 \qquad (2)$$

*Correct Classification Rate:* The Correct Classification Rate noted CCR, also known as Success rate, is the sum of good detections represented by *TP* and *TN* divided by the total number of samples *T*.

$$CCR = \frac{TP + TN}{T} * 100$$

### 4.2.2. Experiments

The objective of our study was to investigate texture classification analysis in an effort to differentiate between normal and pathological subjects of osteoporosis pathology.

Comparative results with percentage of correct classification for our proposed method and classical features extraction methods are summarized in Table 3.

We present in Figure 6, two box plots of the training and classified different simples using SVM classifier.

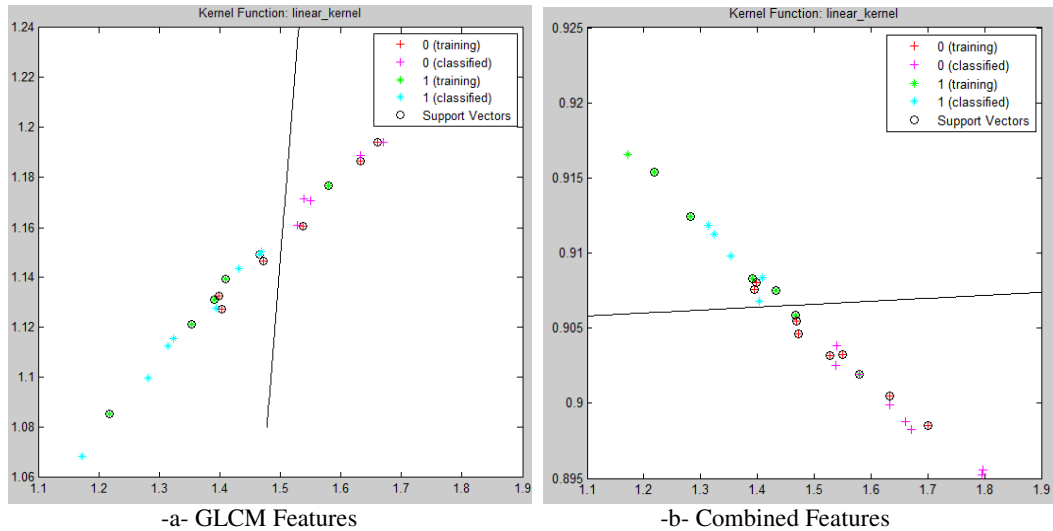

-a- GLCM Features                    -b- Combined Features

Figure 6. The training and classified set of different simples using the SVM classifier

Table 3 shows in percent the confusion matrix of each method, that present the average performance of the SVM classifier for each individual class with the actual class represented on the left and the predicted class represented above. Like this Table, the diagonally shaded boxes show the percent of correctly classified images and the percent of erroneously classified images in the off-diagonal.

Table 3. The average performance of SVM classifier using different methods.

| Features extracted method | Fractal Features | | GLCM Features | | Proposed Method | |
|---|---|---|---|---|---|---|
| | Normal(1) | Abnormal(0) | Normal(1) | Abnormal(0) | Normal(1) | Abnormal(0) |
| Normal (1) | 78.50% | 21.50% | 79.20% | 20.80% | **90.23%** | 9.77% |
| Abnormal (0) | 23.50% | 76.50% | 22.20% | 77.80% | 18.81% | **81.19%** |

Table 4 showed that the classification results using (SVM) classifier of our proposed method succeeded in differentiating between patients. However the CCR is 77.50% (sensitivity, 78.50%; specificity, 76.50%) for fractal Features and 78.50% (sensitivity, 79.20%; specificity, 77.80%) for GLCM features. The best classification results were obtained from including the combination of GLCM Features and Fractal Feature. Also the best percentage of correct classifications achieved was 85.71% (sensitivity, 90.23%; specificity, 81.19%).

Table 4. Percentage classification accuracy (sensitivity; specificity) for each feature extraction method.

| | Fractal Features | GLCM Features | Proposed Method |
|---|---|---|---|
| Sensitivity (%) | 78.50 | 79.20 | 90.23 |
| Specificity (%) | 76.50 | 77.80 | 81.19 |
| CCR (%) | 77.50 | 78.50 | **85.71** |

# 5. CONCLUSIONS AND FUTURE WORK

In this article, we demonstrated the texture classification based on a combination of fractal and GLCM features prove best classification rate. In this method, we propose novel extracted features. The fractal dimension is not sufficient to describe a texture. Therefore a combination with other features will be needed to give good results.

This paper presents a trabecular bone classification method using a combined method based on Fractal and GLCM features. In this method, we propose novel extracted features.

According to experimental results performed on real MRI and CT Scan Medical images, the classification method provides a correct classification rate of 85.71%.

Further research works on a larger number of fractal features is required for validating the results of this study and for finding additional shape and texture features that may provide information for the prevention of osteoporosis on the initial MRI and CT Scan images; and improve the final classification rate. Additionally the use of other classifiers could suggest a better classification a scheme and will be examined.

## ACKNOWLEDGMENT

The authors extend their sincere thanks to Dr. Patrick Dubois, university professor and hospital practitioner in Institute of medical technology, hospital and Regional – Lille, France, for his help about the images database.

## AUTHORS

**Redouan KORCHIYNE** has obtained the Diploma of Advanced Higher Studies "DESA" in Informatics and Telecommunications in 2005 at the Ibn Tofail University Faculty of Science, Kenitra-Morocco. Currently, he is pursuing her PhD degree at the Ibn Tofail University Faculty of Sciences - Kenitra, her research focuses on the texture analysis of medical images by the multifractal approach.

**Sidi Mohamed FARSSI,** Dr. of State es-science, Director of the Doctoral School of Telecommunications at the University of Dakar-Senegal, Professor at the Polytechnic Higher School of Dakar, President of the National Council of Moroccans in Senegal (CNMS) and Director of the laboratory for research in medical imaging and bioinformatics.

**Abderrahmane SBIHI** is a University Professor, currently is a Director of the National School of Applied Sciences, Tangier, Morocco. He is president of the Moroccan association for the development of electronics, computer and Automatic AMDEIA and a member of the standing committee of the African colloquy on Research in Computer Science at Morocco. His main research interests are computer vision, pattern recognition and multidimensional data analysis.

**Rajaa TOUAHNI** Is a Professor at the Ibn Tofail University Faculty of Sciences. Kenitra - Morocco. Responsible for the research team in Information Processing and engineering of the Decision. His main research interests are on image processing and analysis of multidimensional data.

**Mustapha TAHIRI ALAOUI** has obtained the degree of National Doctorate in 2005 at the university Mohammed V, faculty of sciences-Agdal of Rabat-Morocco within the "UFR": Analysis and Design of Computer Systems. His research focuses on the texture characterization of the ultrasound images.